# Cascade Feature Aggregation for Human Pose Estimation

Zhihui Su, Ming Ye, Guohui Zhang, Lei Dai, Jianda Sheng

Nanjing Development Team, Ping An Technology (Shenzhen) Co., Ltd., Nanjing, Jiangsu, China, {suzhihui002, yeming442, zhangguohui128, dailei197, shengjianda720}@pingan.com.cn

**Abstract**

Human pose estimation plays an important role in many computer vision tasks and has been studied for many decades. However, due to complex appearance variations from poses, illuminations, occlusions and low resolutions, it still remains a challenging problem. Taking the advantage of high-level semantic information from deep convolutional neural networks is an effective way to improve the accuracy of human pose estimation. In this paper, we propose a novel Cascade Feature Aggregation (CFA) method, which cascades several hourglass networks for robust human pose estimation. Features from different stages are aggregated to obtain abundant contextual information, leading to robustness to poses, partial occlusions and low resolution. Moreover, results from different stages are fused to further improve the localization accuracy. The extensive experiments on MPII datasets and LIP datasets demonstrate that our proposed CFA outperforms the state-of-the-art and achieves the best performance on the state-of-the-art benchmark MPII.

## 1. Introduction

Human pose estimation refers to the task of recognizing postures by localizing body keypoints from images. It is vital prerequisite step for many computer vision tasks such as human action recognition [33, 34], tracking [35, 36], human-computer interaction [37] and video surveillance [41]. In the past few decades, many efforts are devoted to build robust human pose estimation models under the controlled and uncontrolled setting [40]. The typical methods include pictorial structures models, hierarchical models and non-tree models [42, 43]. The pictorial structures model constructs a classical tree-structured graphical framework by exploring spatial correlations between parts of the body and kinematic priors that couple connected limbs. The hierarchical models represent the relationships between parts at different scales in a hierarchical tree structure, leading to capture high-order relationships among parts and characterize an exponential number of plausible poses [46]. Non-tree models use loops to augment the tree structure with additional edges, which can well capture symmetry, occlusion and long-range relationships. Although the methods mentioned above achieves promising results for human pose estimation under the controlled settings, they usually degenerate severely under the wild scenario due to complex appearance variations from various poses, different illuminations, partial occlusions, etc.

Recently great progresses have been achieved by employing deep neural networks [1, 2, 7, 9, 39, 40]. As an encoder-decoder architecture, hourglass network [7] can well capture information of various scales for robust human pose estimation. Moreover, several hourglass networks are stacked to implement a mechanism for repeated bottom-up, top-down inference allowing for reevaluation of initial estimates and features across the whole image. As an encoder-decoder model, hourglass use a highway to connect the encoder and decoder parts. More detail local information is brought to decoder to improve the performance. Different from stacked hourglass, CPN [39] design another strategy for cascading multiple stages, i.e., a GlobalNet followed by a RefineNet. The GlobalNet aims at obtaining an initial human pose which is slightly different from FPN [44] by applying 1 x 1 convolutional kernel before each element-wise sum procedure in the up-sampling process. The following RefineNet is designed to explicitly address the "hard" keypoints. Besides the top-down methods mentioned above, the bottom-up method like OpenPose [40] can also achieve promising results for real-time human pose estimation.

In this paper, we further push the frontier of the area by resorting to cascade deep networks. We propose a novel Cascade Feature Aggregation (CFA) method to improve the human pose estimation results by leveraging featrues at different stages of the cascade networks, as shown in Fig. 1 Instead of a single stage network, our CFA consists of everal hourglass networks, each figuring out part of the nonlinearity for mapping the input image to the corresponding human body keypoints. Specially, the first stage predicts human poses by taking an input as input and the following stage leverages the featuers of shallow and deep layers from the previous stage to further improve the estimation results. Features from different stages are aggregated to obtain both local detailed information and global context information for robust human pose estimation. Benifited from the advantages of aggregating features from different stages, our CFA is more robust to poses, illuminations and partial occlusions than [1, 2, 6, 9]. Besides, our CFA fuse the results from each stage to further improve the human pose estimation results.

The main contributions of this paper can be summarized as:

1) We proposed a novel Cascade Feature Aggregation (CFA) method for robust human pose estimation by leveraging features from different stages under a cascade structure.

2) By fusing the results from different stages, CFA can further improve the results for human pose estimation.

3) Our CFA outperforms the state-of-the-art methods and achieves the best performance on the MPII benchmark.

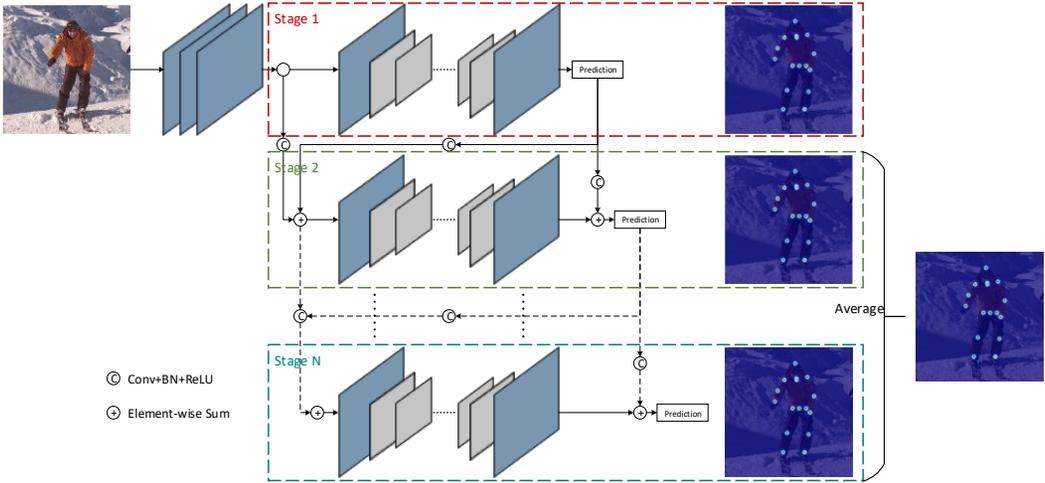

**Figure 1.** The architecture of the proposed CFA. The CFA consists of several stages and each stage is designed to predict the keypoints of the human body. We perform feature aggregation by taking the features from the preview stage as the input for the current stage. Besides, results from multiple stages are fused to further improve the performance.

## 2. Related Work

Human pose estimation is an active research topic for decades. Traditional methods rely on hand-craft features, which formulate the problem of human keypoints estimation as a tree-structured or graphical model problem. Recently, many approaches take advantage of deep convolutional neural network to enhance the performance of pose estimation. In terms of network architecture, current human pose estimation methods could be divided into two categories: single-stage approaches and multi-stage approaches.

**Single-Stage Approach.** The most single-stage approaches concentrate on designing the basic network structure. Hourglass is one of the most typical single-stage approaches. The hourglass considers the information at different scale. High layers have a coherent understanding of the full body, and high way connections bring the essential local evidence to high layers for identifying features like faces and hands. Many researches are devoted to improving the basic network architecture of hourglass. CPN takes the advantage of both hourglass and FPN, leading to the best results on COCO 2017. Xiao et al. simplify the hourglass by removing the high way connection and find four is the best kernel size for deconvolution [11]. Moreover, Tang et al. attempt to add connections and intermediate supervision to avoid potentially large state spaces for higher-level parts, which can overcome some overlapping parts and clutter backgrounds problems [2].

Different from the hourglass and its variant, OpenPose predict the keypoints under a bottom-up pipeline, which first detect the body parts and then joint them up. The algorithm can perform in real-time for human pose estimation on a Nvidia GTX 1080 Ti GPU platform.

**Multi-Stage Approach.** The multi-stage approach focuses on building multiple stages to further improving the performance. CPM [24] firstly introduces multi-stage structure with several convolution and pooling layers, which reaches 88.5% PCKh on MPII test set. CPM illustrates that a sequential architecture composed of convolutional networks is capable of implicitly learning a spatial model for pose by communicating increasingly refined uncertainty-preserving beliefs between stages. Stack hourglass stacks several hourglass networks to achieve more powerful architecture for human pose estimation. For approximately the same number of parameters, the most improvement in final performance for each successive network increases from stacking. Finally, the stacked hourglass improves the PCKh to 90.9% on MPII. Furthermore, several attempts have been made to improve the backbone network for each stage of stacked hourglass. Xiao proposed HRUs to replace the basic high way connections [29], which can enrich the expressive power of conventional residual unit. Xiao et al. also use multi-context attention to guide context modeling, and CRF replace the global softmax for spatial correlation modeling during training. It achieves a further improvement of 91.5% on MPII. Yang et al. propose PRMs to replace the basic convolutions and deconvolutions [42], which enhances the invariance in scales of the DCNNs and improve the PCKh to 92.0%. Ke et al. [2] propose multi-scale supervision network to enhance the supervision, which works hand-in-hand with the structure-aware loss design, to infer high-order structural matching of detected body keypoints, and improve the PCKh to 92.1%. Li et al. employ ResNet as an encode part and propose feature aggregation across stages, which results in the best performance on COCO keypoint challenge 2018.

Generally multi-stage architecture performs better than single-stage methods. And an elaborate design for each stage is also important. Hourglass is a good basic architecture for human pose estimation. The main difference of our CFA from the Belagiannis et al. [22] is that we employ multi-stage structure rather than RNN and induce connections between predictions for better results. Moreover, instead of using large convolution kernels proposed in Belagiannis et al. [22] and Wei et al. [24], we obtain large receptive fields by aggregating features. The key difference between our CFA and Newell [7], Chu [29], Zhang [1] is that the existing methods only take the outputs from the previous stage as the input for the current stage while our CFA takes both the inputs and outputs from the previous stage for the next stage.

## 3. Network Architecture

In this paper, we present a novel Cascade Feature Aggregation (CFA) for robust human pose estimation. Firstly, we will give an overview of our CFA. Secondly, we will describe the backbone model designed for each stage. Thirdly, we will illustrate the details of cascade feature aggregation. Then we will introduce a roust result fusion approach by leveraging heatmaps of several stages. And finally, we will give a detailed discussion about the difference from the existing works.

### 3.1. Overview

As shown in Fig. 1, the proposed CFA attempts to design a general cascade framework in a coarse-to-fine architecture. Specifically, the CFA framework consists of several successive cascaded hourglass with features aggregations. Each stage attempts to characterize the nonlinear mappings from body image to body keypoints in same feature input transformed from the original image.

The first stage endeavors to roughly approximate the body keypoint locations, and therefore an hourglass is designed for predict the keypoints. After getting an estimation of keypoints from the first stage, the successive hourglasses make an effort to refine the shape by combining the preview prediction and current prediction. Furthermore, the final heat map is averaged by the last few stages for further improving the keypoints predictions.

Support we have a training set $\{(x_i, z_i)\}_{i=1}^{N}$, which consists of N face images $x_i$ and its corresponding $p$ body keypoint $z_i \in R^{2p}$. We use gaussian kernel to present the $p$ points in an image with $p$ channels, i.e., the heatmaps for $p$ points, denoted as $y_i$. The human pose estimation task can be reformulated as seeking a mapping function $f : y_i = f(x_i)$

### 3.2. Backbone model

To meets the requirements of capturing various information of different scales, we resort to Hourglass as our backbone model. In the original Hourglass, 4 is chosen as the kernel size of deconvolution [11], which can ensure every point is interpolated by 4 points when conducting up sampling. To improve Hourglass, we employ a deeper network architecture ResNet as backbone model for both encoder and decoder parts. In our experiments, we find that it is useful to make the networks deeper for both the encoder and decoder and is more effective for the encoder. The reason behind it is that encoder with deep network can well capture semantic information at different scales. Several typical ResNets like ResNet-50, ResNet-101 and ResNet-152 are compared and investigated in Sec. 4.3.

For backbone model, the seeking function $f$ can be expressed as $f(x)=f_3(f_2(f_1(x)))$, where $x$ is the input image for prediction, $f_1, f_2, f_3$ is three parts of the network, and $a_i$ is recorded as the result of $f_i$, then we can achieve that $a_1 = f_1(x)$, $a_2 = f_2(a_1)$, $a_3 = f_3(a_2)$, $y = a_3$.

Finally we calculate prediction of the keypoints $z$ by $z = g(y)$, $g$ donates the strategy to calculate the keypoint from feature map. In this paper, we treat the coordinate of max value on the heatmap as the prediction function $g$.

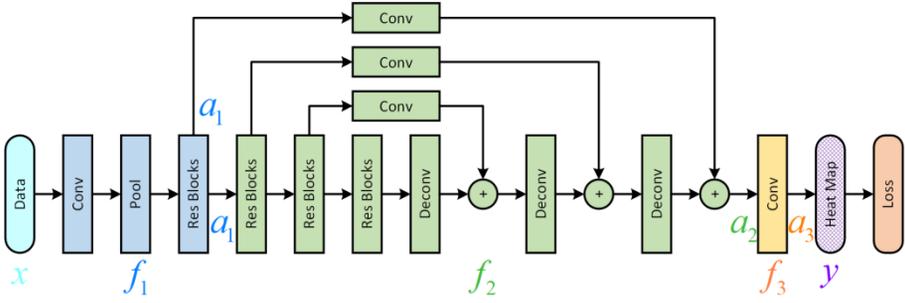

**Figure 2.** ResNet based hourglass network. The network is an encoder-decoder model, which is designed based on hourglass. ResNet is used as the basic structure for the encoder parts. Highway connection is employed from the encoder to the decoder.

### 3.3. Cascaded Feature aggregation

Multi-stage model is designed to enhance the prediction of the invisible keypoints. Suppose we have a training set $\{(x_i, z_i)\}_{i=1}^{N}$, we mapping the ground truth keypoints $z_i$ to the heatmap $y_i^*$ by gaussian function. We calculate the predicted heatmap of each stage by,

$$y_{i,j} = \begin{cases} a_{i,j,3} & j=1 \\ a_{i,j,3} + \varphi_{j,4}(y_{i,j-1}) & j \geq 2 \end{cases} \quad (1)$$

Here, j donates the j-th stage. For the first stage, the network output $a_{i,j,3}$ is learn to approximate the ground truth heatmap $y_i^*$. For the latter stages, a non-linearity function $\varphi_{j,4}$ is designed by taking the heatmap from previous stage as input and the output from $\varphi_{j,4}$ are further added with $a_{i,j,3}$ to achieve the final heatmap estimation. It can amplify points with high confidents and block the miss predicted points.

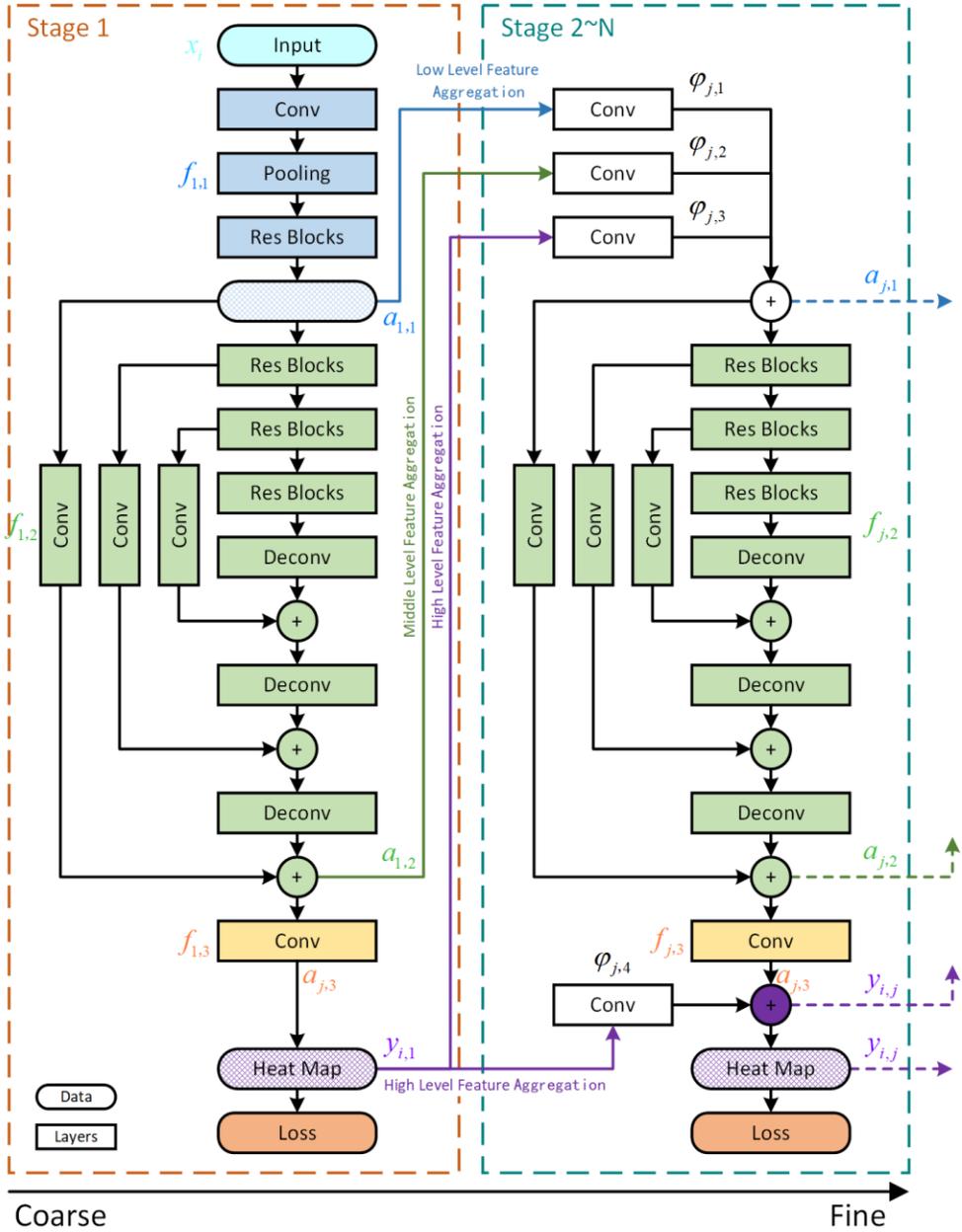

**Figure 3.** Aggregation between different stage of CFA. There are three different aggregation between different stages. The input aggregation brings the local detail information for miss predicted point for a second time prediction. The feature aggregation takes the high layer sematic to input layer. And the prediction aggregation keeps the prediction more stable.

We simplify $a_{i,j,3}$ into $a_{j,3}$ for general samples. For each stage, $a_{j,3}$ is calculated by the following function:

$$a_{j,3} = f_{j,3}\left(f_{j,2}\left(a_{j,1}\right)\right) \tag{2}$$

Here, $a_{j,1}$ is the input of current stage. $f_{j,2}$ and $f_{j,3}$ has the same struct as $f_2$ and $f_3$ respectively in the backbone model mentioned in Sec. 3.2. For the first stage, $a_{1,1}$ is calculated from the original image:

$$a_{1,1} = f_{1,1}(x) \tag{3}$$

For the successive stages, $a_{j,1}$ consists of three parts, i.e., low-level features, middle-level features and high-level features from the previous stage, as shown in Fig. 3. The low-level features contain detailed local information which is beneficial for localizing the exact positions for human parts. While high-level features contain global semantic information, which can improve the performance for partial occlusions and complex backgrounds. Specifically, the $a_{j,1}$ can be achieved by the following equation:

$$a_{j,1} = \varphi_{j,1}\left(a_{j-1,1}\right) + \varphi_{j,2}\left(a_{j-1,2}\right) + \varphi_{j,3}\left(y_{j-1}\right) \qquad j \geq 2 \tag{4}$$

Here $\varphi_{j,1}$, $\varphi_{j,2}$, $\varphi_{j,3}$ donate low level, middle level, high level aggregation functions respectively from the preview stage for the inputs of the current stage.

### 3.4. Results improvement by fusing heatmaps

We fuse the heatmaps of the last few stages to further improve the results. As shown in Fig.4, the prediction is determined by averaging the output heatmaps of last several stages:

$$\sigma_{fusion} = \frac{\sqrt{\sum_{i=N-n}^{N} \sigma_i^2}}{n} \tag{5}$$

Where $\sigma_{N-n} \cdots \sigma_N$ denotes the heatmaps of the last N stages. The fusion strategy makes the results more stable.

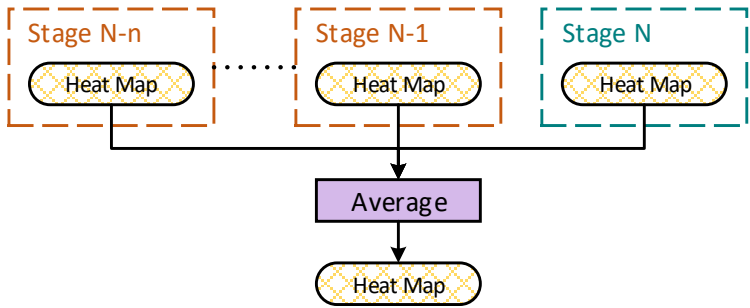

**Figure 4.** Heatmap fusion. The final heat map is averaged by the last several prediction of heat map.

## 3.5. Discussions

**Differences with Li et al. [6].** Both model use hourglass and the coarse-to-fine supervision for different stages. But they differ in the following aspects: Li et al. use cross stage feature aggregation. two separate information flows are introduced from down-sampling and up-sampling units of the previous stage to the down-sampling procedure of the current stage for each scale. But we use low level, middle level and high-level aggregation as the input of next stage. We think different features from different stages are aggregated to obtain both local detailed information and global context information for robust human pose estimation. The high-level global context can highlight the important local parts, which can help refine the predictions. On the other hand, the low-level original detail information can help re-predict the miss predicted keypoints based on the global information.

**Differences with Zhang et al. [1].** Both Zhang et al. [1] and our CFA use hourglass for each stage and prediction fusion with prediction aggregated from preview stage. But they differ in the following aspects: 1) Zhang et al. [1] use a PGNN for further enhancing the accuracy. They believe that the contextual information propagated from confident parts through graph helps reducing error. Differently, we cascade several stages to get high-level semantic features which can capture both global semantic information and local detailed information, leading to better results. 2) Zhang et al. use the ResNet-50 as backbone model, the high layers feature map of preview stage as the input of current stage. We also use ResNet as the backbone model, and ResNet-101 is suggested for first stage. ResNet has four residual blocks, only the last three blocks are used in the following stages, and the first stage is shared by input aggregation.

# 4. Experiments

In this section, we firstly illustrate the experimental settings for the evaluations including the datasets and metric for comparison; secondly explanation the training details; thirdly investigation of our CFA; then presenting the improvement of results fusion and inference of training with additional data; finally compare the proposed CFA with the state-of-the-art methods.

We train and test our model on a server with 8 NVIDIA Tesla V100 16G GPUs. All the experiments are doing on the PyTorch 1.0 with NVIDIA apex. All the batch normal used in the net is cross-GPU synchronized batch normal. The batch size is 64 for every experiment. The input size of image is 384x384, multi augmentation is used, like rotation, flipping, scaling, brightness, contrast, saturation, hue. The start of learning rate is 5e-4 and the decay weight is 0.3 for epoch 6, 10, 13. The gradient scheme is Adam.

## 4.1. Dataset and evaluation metric

To evaluate the effectiveness of the proposed CFA algorithm, two public datasets are used for our experiments, MPII [12], LIP [45]. MPII dataset includes around 25K images containing over 40K people with annotated body joints. the dataset was systematically collected and covers 410 every day human activities. LIP dataset contains 50K images with elaborated pixel-wise annotations with 19 semantic human part labels and 2D human poses with 16 keypoints. The images collected from the real-world scenarios contain human appearing with challenging poses and views, heavily occlusions, various appearances and low-resolutions.

We compare our method with a few state-of-the-art methods, i.e., DLCM [2], Adversarial PoseNet [31], MCA [29]. The standard PCKh metric [11] is employed for evaluation. All the performance listed below is showed in PCKh @ 0.5, which means PCK measure that uses the matching threshold as 50% of the head segment length. Moreover, the double flipping test method is utilized for better performance, which calculates the final heat map by averaging the heat maps of the original image and the flipped image. The final predictions are determined by selecting the point from the maximum response of the multi-stage heat maps.

## 4.2. Training details

When conducting experiments on MPII, two kind of training sets are used, i.e., only the MPII training dataset, and both the MPII training dataset and HSSK (Human Skeletal System Keypoints [10]). The first stage of CFA is initialized from the model trained on ImageNet and the parameters of other stages are randomly initialized. In our experiments, it is hard for CFA of more than three stages to achieve convergence with random initializations, which means you should firstly train a triple-stages CFA and then employ the triple-stages CFA model for initializations for leaning a quad-stages CFA model. For experiments on LIP, all the models are initialized for the models trained on MPII.

### 4.3. Investigation of our CFA

Since our CFA consists of multi-stages, we investigate different ResNet architectures as backbone models for each stage. Firstly, we evaluate different backbone models, i.e., ResNet-50, ResNet-101 and ResNet-152 for the first stage. Table 1 shows the prediction results of stage one for human pose estimation. MPII training data is used for training. As seen, ResNet-101 performs better than ResNet-50 and ResNet-152 performs slightly better than ResNet-101. The performance is gradually improved with deeper and deeper networks. In consideration of both accuracy and computation cost, ResNet-101 is chosen for the first stage.

Table 1. Performance of backbone models

| The first Stage | PCKh @ 0.5 |
|---|---|
| ResNet-50 | 88.48 |
| ResNet-101 | 89.26 |
| ResNet-152 | 89.59 |

We further exploit the network architectures for the following stages. In our experiments, we fix the first stage with ResNet-101 and evaluate the successive stages with ResNet-50 and ResNet-101 respectively. Results of different stages are shown in Table 2. ResNet-50 achieves the comparable results with ResNet-101. It can be concluded that RestNet-50 is deep enough for the following stages.

Table 2. Compared between different successive ResNet (PCKh @ 0.5)

| Net | Double-Stage | Triple-Stage | Quad-Stage |
|---|---|---|---|
| Cascaded with ResNet-50 | 89.72 | 89.95 | 89.95 |
| Cascaded with ResNet-101 | 89.73 | 89.94 | 89.87 |

After determining the backbone models for each stage, we explore how many stages of CFA is enough for human pose estimation. We employ ResNet-101 for the first stage and ResNet-50 for the other stages. The model is trained on MPII dataset. As shown in Fig. 5, the performance is increased gradually with more and more stages. After cascading four stages, the performance is keeping stable.

Besides, more training data may need more stages to well capture the complex non-linearity from images to human poses. So, we do a series of experiments to seek how many stages can achieve the performance peak with more training data. The model is trained on both the MPII and HSSK datasets. As seen in Fig. 5, cascading one more stage can achieve better result and additional training data is helpful for improving final results. The influence of additional training data is detailly discussed in the following section. In the following experiments, we all takes ResNet-101 for the first stage and ResNet-50 for the last stages.

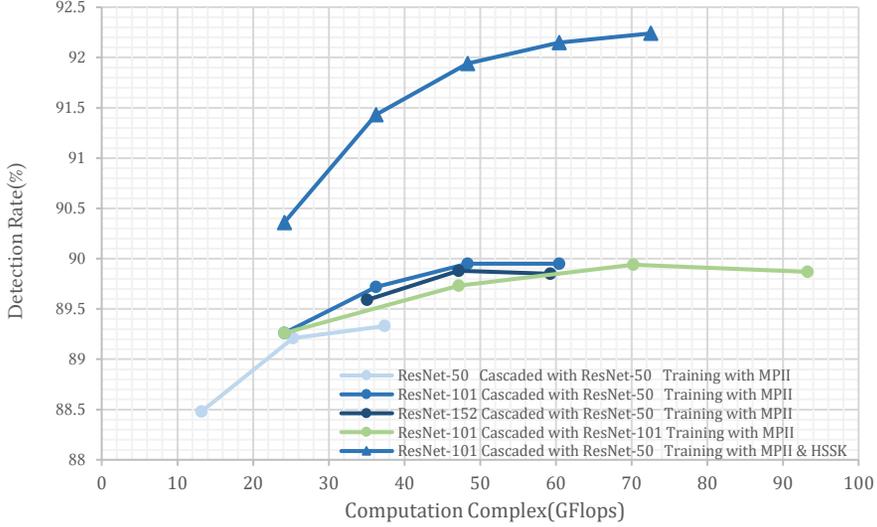

**Figure 5.** Comparison of different multi-stage models (PCKh @ 0.5). For each line from left to right, for example ResNet-101 cascaded with ResNet-50, the first point means single stage of ResNet-101, the second point means a double-stages model constructed from ResNet-101 and cascaded with ResNet-50, and so on.

### 4.4. Comprehensive evaluations for each stage and results fusion

In this section, we give the comprehensive evaluations for each stage and the results fusion. As shown in Table 3, CFA with more and more stages performs better and better. Moreover, for CFA with quad-stages, the performance gradually improves from 94.35 at stage 1 to 95.82 at stage 4, which demonstrate the effectiveness of cascade feature aggregation. Benefited from the results fusing, the human pose estimation result can be further improved to 95.85.

Table 3. Intermediate and fused predictions for multi-stage model
(MPII Validation, PCKh @ 0.5)

| Net | Stage | Head | Shoulder | Elbow | Wrist | Hip | Knee | Ankle | Total |
|---|---|---|---|---|---|---|---|---|---|
| Single-Stage | 1 | 95.49 | 95.33 | 90.29 | 85.81 | 88.74 | 86.49 | 82.01 | 89.26 |
| Double-Stages | 1 | 95.32 | 95.00 | 90.35 | 86.04 | 89.48 | 87.17 | 81.92 | 89.41 |
| Double-Stages | 2 | 95.43 | 95.05 | 90.59 | 86.52 | 89.62 | 87.69 | 82.51 | 89.72 |
| Triple-Stages | 1 | 94.97 | 95.55 | 90.87 | 86.29 | 88.14 | 87.07 | 82.66 | 89.52 |
| Triple-Stages | 2 | 95.31 | 95.65 | 91.07 | 86.51 | 88.35 | 87.35 | 83.28 | 89.77 |
| Triple-Stages | 3 | 96.00 | 95.65 | 91.07 | 86.46 | 89.25 | 87.09 | 83.63 | 89.95 |
| Triple-Stages | fused | 96.13 | 95.72 | 91.33 | 86.40 | 89.25 | 87.53 | 83.61 | 90.06 |
| Quad-Stages | 1 | 94.35 | 95.44 | 90.81 | 86.64 | 88.78 | 87.06 | 82.35 | 89.53 |
| Quad-Stages | 2 | 94.69 | 95.46 | 91.10 | 86.81 | 88.7 | 87.63 | 83.53 | 89.87 |

| | | | | | | | | |
|---|---|---|---|---|---|---|---|---|
| Quad-Stages | 3 | 95.07 | 95.44 | 91.10 | 86.95 | 88.66 | 87.57 | 83.70 | 89.92 |
| Quad-Stages | 4 | 95.82 | 95.37 | 90.83 | 86.81 | 89.56 | 87.23 | 83.46 | 89.95 |
| Quad-Stages | fused | 95.85 | 95.37 | 90.96 | 86.93 | 89.77 | 87.63 | 83.91 | 90.14 |

We can easily find that the lower stage's performance will be increased with the depth of the stage.

### 4.5. Training with additional data

To exploit how training data contributes to the performance improvement for human pose estimation, we conduct experiments on MPII dataset and HSSK database. As shown in Table 4, the model trained on only MPII dataset achieves 89.95 PCKh @ 0.5, and the performance keeps stable with four stages. After training with additional dataset HSSK, the performance is significantly improved to 92.24 and cascading more stages, i.e., five stages can achieve better results. The additional data is crucial for improving performance.

Table 4. CFA trained on different datasets (PCKh @ 0.5)

| Data | Single-Stage | Double-Stage | Triple-Stage | Quad-Stage | Five-Stage |
|---|---|---|---|---|---|
| MPII Dataset | 89.26 | 89.72 | 89.95 | 89.95 | - |
| MPII & HSSK Datasets | 90.36 | 91.43 | 91.94 | 92.15 | 92.24 |

### 4.6. Comparison with the Existing Methods

Table 5. Comparison with the existing methods on MPII test set (PCKh @ 0.5)

| Method | Head | Shoulder | Elbow | Wrist | Hip | Knee | Ankle | PCKh |
|---|---|---|---|---|---|---|---|---|
| Pishchulin et al. [13] | 74.3 | 49.0 | 40.8 | 34.1 | 36.5 | 34.4 | 35.2 | 44.1 |
| Tompson et al. [14] | 95.8 | 90.3 | 80.5 | 74.3 | 77.6 | 69.7 | 62.8 | 79.6 |
| Carreira et al. [15] | 95.7 | 91.7 | 81.7 | 72.4 | 82.8 | 73.2 | 66.4 | 81.3 |
| Tompson et al. [16] | 96.1 | 91.9 | 83.9 | 77.8 | 80.9 | 72.3 | 64.8 | 82.0 |
| Hu & Ramanan. [17] | 95.0 | 91.6 | 83.0 | 76.6 | 81.9 | 74.5 | 69.5 | 82.4 |
| Pishchulin et al. [18] | 94.1 | 90.2 | 83.4 | 77.3 | 82.6 | 75.7 | 68.6 | 82.4 |
| Lifshitz et al. [19] | 97.8 | 93.3 | 85.7 | 80.4 | 85.3 | 76.6 | 70.2 | 85.0 |
| Gkioxary et al. [20] | 96.2 | 93.1 | 86.7 | 82.1 | 85.2 | 81.4 | 74.1 | 86.1 |
| Rafi et al. [21] | 97.2 | 93.9 | 86.4 | 81.3 | 86.8 | 80.6 | 73.4 | 86.3 |
| Belagiannis & Zisserman [22] | 97.7 | 95.0 | 88.2 | 83.0 | 87.9 | 82.6 | 78.4 | 88.1 |
| Insafutdinov et al. [23] | 96.8 | 95.2 | 89.3 | 84.4 | 88.4 | 83.4 | 78.0 | 88.5 |
| Wei et al. [24] | 97.8 | 95.0 | 88.7 | 84.0 | 88.4 | 82.8 | 79.4 | 88.5 |
| Bulat & Tzimiropoulos [25] | 97.9 | 95.1 | 89.9 | 85.3 | 89.4 | 85.7 | 81.7 | 89.7 |
| Newell et al. [7] | 98.2 | 96.3 | 91.2 | 87.1 | 90.1 | 87.4 | 83.6 | 90.9 |
| Tang et al. [26] | 97.4 | 96.4 | 92.1 | 87.7 | 90.2 | 87.7 | 84.3 | 91.2 |
| Ning et al. [27] | 98.1 | 96.3 | 92.2 | 87.8 | 90.6 | 87.6 | 82.7 | 91.2 |
| Luvizon et al. [28] | 98.1 | 96.6 | 92.0 | 87.5 | 90.6 | 88.0 | 82.7 | 91.2 |
| Chu et al. [29] | 98.5 | 96.3 | 91.9 | 88.1 | 90.6 | 88.0 | 85.0 | 91.5 |
| Chou et al. [30] | 98.2 | 96.8 | 92.2 | 88.0 | 91.3 | 89.1 | 84.9 | 91.8 |
| Chen et al. [31] | 98.1 | 96.5 | 92.5 | 88.5 | 90.2 | 89.6 | 86.0 | 91.9 |
| Yang et al. [32] | 98.5 | 96.7 | 92.5 | 88.7 | 91.1 | 88.6 | 86.0 | 92.0 |
| Ke et al. [9] | 98.5 | 96.8 | 92.7 | 88.4 | 90.6 | 89.3 | 86.3 | 92.1 |
| Tang et al. [2] | 98.4 | 96.9 | 92.6 | 88.7 | 91.8 | 89.4 | 86.2 | 92.3 |
| Zhang et al. [1] | 98.6 | 97.0 | 92.8 | 88.8 | 91.7 | 89.8 | 86.6 | 92.5 |
| Ours Triple-Stage Model | 98.1 | 97.2 | 93.4 | 89.1 | 91.2 | 90.1 | 86.7 | 92.6 |
| Ours Five-Stage Model | **98.7** | **97.5** | **94.3** | **90.7** | **93.4** | **92.2** | **88.4** | **93.9** |

Firstly, we evaluate our CFA with the existing methods on MPII. Ke et al. stack the stages by using high-level feature map from preview stage and achieve 92.1 PCKh @ 0.5 [9]. Zhang et al. [1] further use the relation between keypoints to improve performance and achieve better results than Ke et al. [9]. Our CFA performs better than Zhang et al. [1] and achieves the best results of 93.3 PCKh @ 0.5 on the MPII, which attributes the cascade feature aggregation and results fusion. Fig. 6 shows the visualization results of our CFA for the first stage and the last stage. As seen, the results of the first stage may fail when two people intersect or body parts are partially occluded. Thanks to the global semantic features from the last stage, the results can be significantly improved and CFA nearly achieves the perfect results for the hard cases mentioned above.

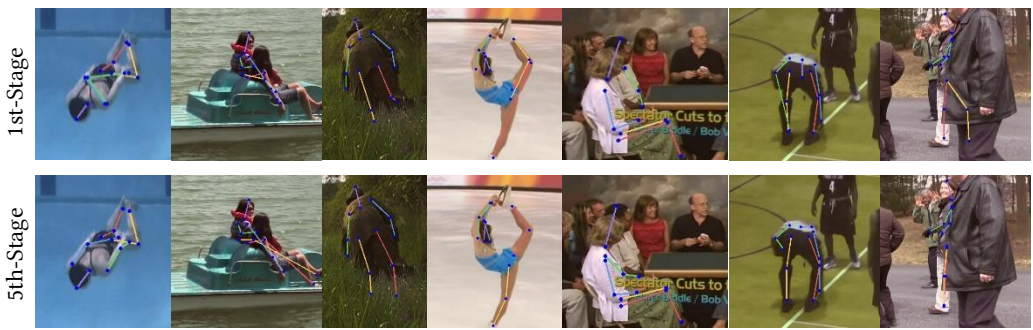

Figure 6. The visualization results of CFA for the first and last stages.

Table 6. Comparison with the existing methods on LIP test set (PCKh @ 0.5)

| Method | Head | Shoulder | Elbow | Wrist | Hip | Knee | Ankle | UBody | Total |
|---|---|---|---|---|---|---|---|---|---|
| Hybrid Pose Machine, Liao et al. | 71.7 | 87.1 | 82.3 | 78.2 | 69.2 | 77.0 | 73.5 | 79.8 | 77.2 |
| BUPTMM-POSE, Liu et al. | 90.4 | 87.3 | 81.9 | 78.8 | 68.5 | 75.3 | 75.8 | 84.8 | 80.2 |
| anonymous | 93.9 | 88.6 | 82.2 | 78.3 | 69.7 | 73.3 | 73.2 | 86.0 | 80.6 |
| Tightly Connected ResNet-101 | 93.2 | 87.9 | 82.3 | 79.8 | 69.1 | 75.5 | 75.4 | 85.9 | 81.0 |
| anonymous | 92.8 | 88.5 | 83.4 | 82.0 | 67.1 | 76.4 | 76.3 | 86.8 | 81.4 |
| anonymous | 93.1 | 89.6 | 84.5 | 82.4 | 69.8 | 77.3 | 77.7 | 87.5 | 82.5 |
| MPP, Tao et al. | 93.6 | 89.7 | 84.5 | 82.8 | 70.1 | 77.0 | 78.5 | 87.8 | 82.8 |
| anonymous | 93.6 | 89.8 | 85.0 | 83.7 | 70.4 | 78.8 | 78.9 | 88.1 | 83.3 |
| anonymous | 94.2 | 91.9 | 87.5 | 84.2 | 75.0 | 83.4 | 83.0 | 89.6 | 85.9 |
| Zll | 94.3 | 92.1 | 87.8 | 85.3 | 75.3 | 82.8 | 82.3 | 90.0 | 86.0 |
| anonymous | 94.2 | 92.4 | 88.5 | 86.1 | 76.0 | 83.6 | 82.5 | 90.4 | 86.5 |
| cs_ft-2, Li et al. | 94.4 | 92.4 | 88.3 | 86.0 | 76.1 | 84.7 | 85.0 | 90.4 | 87.0 |
| LTL | 94.9 | 93.1 | 89.9 | 87.6 | 75.9 | 84.9 | 84.4 | 91.4 | 87.5 |
| EP108 | 95.0 | 93.0 | 89.1 | 86.7 | 75.9 | 85.5 | 86.2 | 91.1 | 87.6 |
| IR-new | 94.1 | 93.1 | 88.4 | 86.5 | 78.0 | 85.9 | 86.0 | 90.6 | 87.6 |
| ByteDance-SEU-Baseline, Su et al. | 95.8 | 94.4 | 91.7 | 89.6 | 80.2 | 89.5 | 89.2 | 93.0 | 90.2 |
| JDAI-human, Li et al. | **95.9** | **94.8** | 92.3 | 90.4 | 81.4 | 90.3 | 90.2 | 93.4 | 90.9 |
| Ours Five-Stage Model | 95.8 | 94.7 | **92.5** | **90.4** | **81.7** | **90.6** | **90.5** | **93.4** | **91.0** |

Furthermore, we conduct experiments on LIP dataset. The comparison results are shown in Table 6. As seen, the similar conclusion can be obtained that our CFA achieves the best results, which demonstrate the effectiveness of capturing both local detailed information and global semantic information by cascade feature aggregation.

Fig. 7 shows some failure cases of our method. As seen, the performance degrades on some images which has complex illumination, low resolution and motion blur, partially due to the lack of such samples in training set.

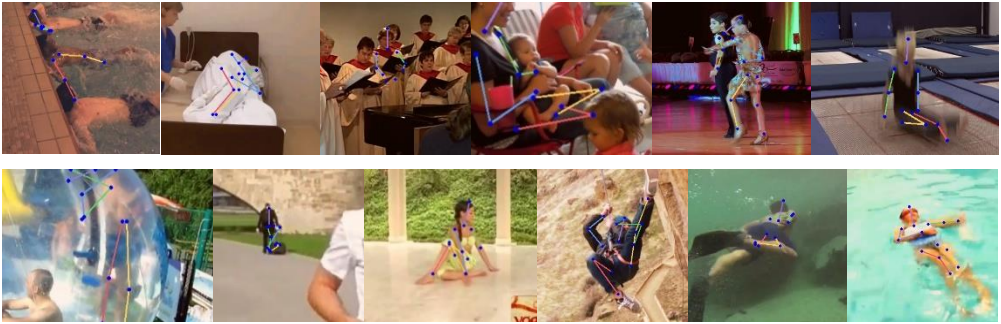

Figure 7. The visualization results of some failure cases.

## 5. Conclusions

In this paper, we propose a novel method CFA for robust human pose estimation, which cascades several hourglasses and aggregates features of low-level, middle-level and high-level to well capture local detailed information and global semantic information. Moreover, the proposed CFA exploits ResNet-101 and ResNet-50 for the first stage and the following stages respectively, which achieves a good trade-off of both accuracy and efficiency. The CFA achieves better results than the state-of-the-art methods like Ke et al. [9] and Zhang et al. [1]. on two datasets MPII and LIP. Besides, the experiments results show that the data diversity is extremely important for improving performance.